\documentclass[11pt,a4paper]{article}
\usepackage[hyperref]{acl2020}
\usepackage{times}
\usepackage{latexsym}

\usepackage{microtype}

\usepackage{url}
\usepackage{graphicx}
\usepackage{multirow}
\usepackage{amsthm}
\usepackage{amssymb}
\usepackage{amsmath}
\usepackage{amsfonts}
\usepackage{xcolor}
\usepackage{booktabs}
\usepackage{subcaption}
\usepackage[noalgohanging]{algorithm2e}

\usepackage{xcolor}
\usepackage{soul}
\usepackage{url}

\aclfinalcopy % Uncomment this line for the final submission
\title{Low Resource Multi-Task Sequence Tagging - Revisiting Dynamic Conditional Random Fields}

\author{Jonas Pfeiffer, Edwin Simpson\thanks{{ } Edwin is now affiliated with the University of Bristol.}, Iryna Gurevych  \\
Ubiquitous Knowledge Processing Lab (UKP) \\
Department of Computer Science, Technische Universit\"at Darmstadt\\
\texttt{pfeiffer@ukp.tu-darmstadt.de} \\
}

\date{}

\begin{document}
\maketitle
\begin{abstract}
%   There has been a recent surge in utilizing Multi-Task Learning (MTL) architectures on a variety of tasks. 
  We compare different  models for
  %The focus of this paper is on 
  low resource \emph{multi-task sequence tagging} that leverage 
  dependencies between label sequences for different tasks. 
  Our analysis is aimed at datasets where each example has
  labels for multiple tasks.
  % 'available' is only true at training time; at prediction time the labels exist but are unknown?
% that contain labels for \textit{all} tasks for each example.
%   a subcategory of multi-task learning wherein the labels of all tasks are available for each data point
%   during training.
  %wherein a sequence of data points corresponds to multiple sequences of labels, each belonging to a different task.
  Current approaches use either a separate model for each task
  or standard multi-task learning
  to learn shared feature representations. % across tasks.
  However, these approaches ignore correlations between label sequences,
  which can provide important information in settings with small training datasets.
  To analyze which scenarios can profit from modeling dependencies between labels in different tasks,
%   To address this, 
  we revisit dynamic conditional random fields (CRFs) and combine them
  with deep neural networks.
%   to model dependencies between labels in different tasks. 
  % The commented text below could be reused in section 5 as an overview of the methods we test?
%   We compare three stages of dependency for Multi-Label Sequence Learning. \textit{First}, in single task setups where a model is trained for each task separately. \textit{Second}, in Multi-Task setups where a feature function such as a BiLSTM is shared between the tasks with independent prediction-heads given the features, in the form of Linear-Chain CRFs. \textit{Third}, Dynamic Conditional Random Field setups, in which we introduce dependency assumptions between the task predictions. For this we revisit both factorial CRFs and cascaded factorial CRFs which we integrate with deep neural networks. 
  We compare single-task, multi-task and dynamic CRF setups for three diverse datasets at both sentence and document levels in English and German %for which we simulate 
  low resource scenarios. We show that including silver labels from pretrained part-of-speech taggers
  as auxiliary tasks can improve performance on downstream tasks. 
  We find that especially in low-resource scenarios, the explicit modeling of inter-dependencies between task predictions outperforms single-task as well as 
  standard multi-task models. 
%   The more data 
%   By explicitly modeling the inter-dependencies between task predictions, %factorial 
%   dynamic CRFs outperform single-task as well as 
%   standard multi-task models
%   across all tasks in our low resource settings.   
  
%   and which has been receiving diminishing attention in recent times. 
% We compare Multi-Task setups in which feature functions, such as BiLSTMs, are shared among the tasks 
% Adapting tasks for which multiple labels exist to Multi-Task techniques in which only the  results in information loss due to independence assumptions. In our work we propose architectural changes to mitigate this information loss by explicitly modeling the inter-dependencies.

%   We investigate the loss of information that results from adapting existing MTL techniques for MLSL tasks, and suggests architectural changes to mitigate it by explicitly modeling the inter-dependencies between different tasks. 
%   We revisit multiple Sequence Labeling (SL) approaches and adapt them to neural settings. 
%   We find that in preliminary results, the additional information helps in generalization, especially for infrequent classes. 
%   The general architectures of MTL alow an easy adaptation to MSLS tasks, however a lot of information is lost when defining this task to be a subcategory of MTL. 
\end{abstract}

\section{Introduction}
% Recent work on Sequence Labeling has primarily focused on different architectures for sequential representation learning. This has been in the form of Convolutional Neural Networks (CNNs), Bidirectional LSTMs \cite{hochreiter1997long} (BiLSTMs) or more recently Transformers \cite{vaswani2017attention}. 

% State the importance of seq labeling; state the problem - these models require large datasets to learn each task separately. 
% some of these details should move to related work? 

We consider the problem of 
\emph{multi-task sequence tagging (MTST)} with small training datasets,
where each token in a sequence has multiple labels, each corresponding to a different task.
Many advances in sequence labeling for NLP stem from 
combining new types of deep neural network (DNN) 
with conditional random fields \cite[CRFs;][]{lafferty2001conditional, sutton2012introduction}.
In these approaches, such as \citet{huang2015bidirectional},
\citet{lample2016neural} and \citet{ma2016end},
DNNs extract rich vector representations from raw text sequences that 
facilitate classification, while
CRFs capture the dependencies between labels in a sequence.
%ES: this actually doesn't fit here
%Recent approaches for representation learning, such as BERT~\cite{devlin2018bert},
%leverage pretrained language models with 
%hundreds of millions of parameters and fine-tune them to a target task.
However, as the DNNs contain many parameters,
strong performance is achieved by training on large labeled 
datasets, %~\cite{something_that_shows_big_data_fine_tuning}, 
which are unavailable for many domain-specific span annotation tasks.

\begin{figure}
  \centering
  \includegraphics[width= .9\linewidth]{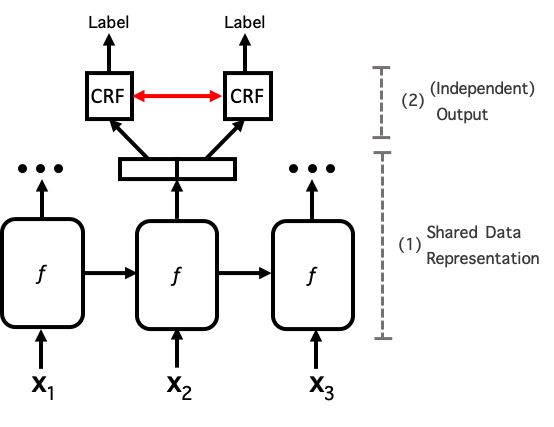}
  \caption{General setup of multi-task sequence tagging (MTST)
  for a sequence of tokens $x_1$ to $x_3$.
  Standard MTL does not model the flow of inter-dependency information  (red arrow) between the output layers (i.e., CRFs). }
  \label{fig:MLSL}
  \vspace{-1.5em}
\end{figure}

%However, given multiple sequence tagging tasks,
The performance of large DNNs can often be improved by 
multi-task learning (MTL), which trains a shared data representation for several related 
tasks so that the representation is learned
from a larger pool of data~\cite{collobert2008unified}.
While this suggests MTL may be a solution for multi-task sequence tagging, 
% for classes that have extremely sparse labels in the training set
% an MTL setup is still unable to perform well.
% MTL from learning suitable representations for those tasks
% as the DNNs with many parameters are unable to l dependencies between tasks from such training data.
standard MTL assumes that the sequences of labels for each task are conditionally independent
given the shared data representations, as depicted in Figure \ref{fig:MLSL}.
These multi-task setups therefore do not model the dependencies between CRFs, illustrated by the red arrow. 
This modeling decision may result in %an unnecessary 
information loss
if important dependencies between tasks are not adequately modeled by the shared representations alone.
Nonetheless, the go-to strategy in recent works is to either tackle the tasks separately
without MTL~\cite{lee2017ukp}, or employ an MTL setup with 
multiple independent linear-chain CRFs in the prediction layer~\cite{Schulz2019aaai}.

As an alternative to the standard CRF, \emph{dynamic CRFs}, such as 
the \emph{factorial CRF}~\cite{sutton2007}, explicitly model dependencies between multiple sequences of labels, but have not previously been integrated with DNNs, so until now have 
relied on fixed text representations that cannot be improved through training.
In this work, we %propose three new approaches to MTST that 
adopt factorial CRFs into a neural setting, finding that 
especially for difficult tasks and low resource settings,
modeling task inter-dependencies outperforms both single task and multi-task setups that do not model the inter-dependencies, 
indicating that this additional flow of information  helps performance considerably. 

Our core contributions are:
(1) a review of different CRF architectures for multi-task sequence tagging (MTST) in a neural network setting;
(2) three new MTST models that integrate factorial CRFs with deep neural networks to exploit dependencies between tasks;
and 
(3) an empirical analysis of different CRF architectures, 
% showing situations where multi-task
% and factorial CRF approaches are most suitable.
showing situations where factorial CRF approaches are more suitable than traditional multi-task learning or single-task setups.

Our implementation extends the popular sequence labeling framework FLAIR\footnote{\href{https://github.com/zalandoresearch/flair}{https://github.com/zalandoresearch/flair}} \cite{AkbikBBRSV19}.
To make future experiments and reproducibility easy, our
experiments use existing publicly available datasets
and we make our code available under \href{https://github.com/UKPLab/multi-task-sequence-tagging}{https://github.com/UKPLab/multi-task-sequence-tagging}.

% \paragraph{Contributions}
% In this paper we revisit different multi-label sequence labeling approaches and lift them into the neural setting. While in cases of sufficient data, Multi-Task learning performs on par with more complex architectures (e.g. factorial CRFs), we show that especially hard tasks with limited data profit from modeling the inter-dependencies between labels. 

\section{Related Work}\label{sec:related_work}

% \subsection{Sequence Learning} 
In recent years, research into sequence labeling has focused
on representations of the input text.
Several architectures were introduced that 
combine word and character embeddings as inputs to a DNN,
evolving from the BiLSTM-CRF~\cite{huang2015bidirectional}
to BiLSTM-LSTM-CRF~\cite{lample2016neural} and BiLSTM-CNN-CRF~\cite{ma2016end}.
% Deep Neural Network (DNN) architectures such as Recurrent Neural Networks (RNNs) in the form of LSTMs \cite{hochreiter1997long} and Convolutional Neural Networks (CNNs) in combination with Conditional Random Fields (CRFs) \cite{lafferty2001conditional, sutton2012introduction} \cite{huang2015bidirectional, lample2016neural, ma2016end}. These architectures combine word and character embeddings as input to the DNN. 
These approaches have been enhanced by leveraging pretrained language models \cite{peters2017semi, liu2018empower} or using contextual embedding representations \cite{akbik2018} such as ELMO \cite{peters2018deep} or BERT \cite{devlin2018bert}. 
However, all of these approaches focus on the data representation
%introducing new feature function on the token level in the form of deep and heavily pretrained neural models, they all leverage
and use a linear-chain CRF as the prediction head, 
so do not model task dependencies in an MTST scenario.
%This has become a default setup as significant performance boosts have been accomplished for linguistic sequence labeling tasks. 

% \subsection{Multi-Task Learning}
% rephrase to tell us what type of problem mtl solved. 
Multi-task learning (MTL) has been widely used in NLP to
%has a long standing history in NLP where especially with the introduction of neural networks many works have focused on leveraging
exploit multiple datasets for representation learning \cite{collobert2008unified, liu2016recurrent, nam2014large, liu2017adversarial}. 
The general architecture of MTL systems consists of two components: (1) a shared data representation, and (2) an (independent) task specific output or prediction layer~\cite{caruana1997multitask, collobert2008unified, nam2014large, liu2016recurrent,  liu2017adversarial, zhang2017survey, ruder2017, ruder2019latent, sanh2019hierarchical}. %change from 'output component' to 'output or prediction layer' for consistency
\citet{sgaard2016} show that in MTL setups, different tasks perform better if the 
prediction layer is on different layers of multi-layer LSTMs for part-of-speech tagging 
(POS), syntactic chunking and CCG supertagging. 
\citet{bingel2017identifying} provide an in depth ablation study on 
which task combinations, 
such as POS, multi-word expressions, super-sense tagging, etc., profit from one another, 
while
\citet{ChangpinyoHS18} design different strategies for sharing weights between tasks. 
More recently, \citet{simpson2020} use variational inference to combine the predictions of multiple taggers trained on different tasks. 
\citet{greenberg2018marginal} train a single CRF from multiple datasets 
using marginal likelihood training to mitigate missing labels.
However, these design traits are necessary because 
each task has different data with 
%the data from different tasks contains
task-specific idiosyncrasies that require different encodings. 
%In order to eliminate these necessary design traits and in order to be able to model task inter-dependencies, 
We eliminate the need for such design traits by focusing on MTST settings
where multiple labels are provided for 
the same set of sentences, thus all tasks share a single data representation.
% ES we have not introduced the idea of feature functions yet.
%only explore architectures where 
%all weights of a feature function $f$ are shared.  

At first glance, the MTST setup seems closely related to Nested NER 
(NNER)~\cite{alex2007recognising, finkel2009nested},
which introduces a hierarchical structure of dependent entities.
However, NNER does not necessarily focus on \textit{different} tasks at the hierarchical level, but allows the same label from the same task
to be tagged over the same span multiple times. 
This is significantly different to MTST, where all overlapping spans correspond
to distinct tasks.  
% Much recent work has focused on designing task specific architectures for NNER. \citet{ju2018neural} model the hierarchical structure of NNER by predicting the lower level hierarchical entities on lower levels of a BiLSTM. \citet{lin2019sequence} target NNER by predicting Anchor words of entities and subsequently predicting regions similar to pointer networks. \citet{luan2019general} introduce a model that leverages dynamic span graphs that, through multiple steps, combines co-reference resolution and entity extraction. \citet{li2019unified} interpret NNER as the task of machine reading comprehension in order to mitigate the effect of overlapping labels. 

In summary, existing work focuses on 
task specific solutions that either 
model a specific hierarchy in NNER~\cite{alex2007recognising,  finkel2009nested, ju2018neural, lin2019sequence, luan2019general, li2019unified},
assume independence between tasks given the shared data
representation~\cite{sgaard2016, bingel2017identifying, greenberg2018marginal, ChangpinyoHS18},
or combine predictions from completely independent taggers~\cite{simpson2020}. 
In contrast, 
we focus on simple CRF procedures that do not require task-specific adaptations,
can be integrated as a prediction layer with any underlying data representation model,
and learn jointly from multiple labels for the same sequence.

% Different strategies for shared weights between tasks can be defined in MTL settings \citet{ChangpinyoHS18}, however, these design traits are primarily necessary because the data from different tasks presumably consist of task specific idosyncracies, which require different encodings. In order to eliminate this possible effect, we focus on datasets where multiple labels are provided for sentences and we thus only explore architectures with shared $f$ encodings for each task.  

\section{Multi-Task Sequence Tagging }
\label{sec:MTST}
    
\begin{figure*}[htp]
\centering
\subcaptionbox{Linear-Chain CRF\label{fig:LCCRF}}[0.2\textwidth]{\includegraphics[width=0.16\textwidth]{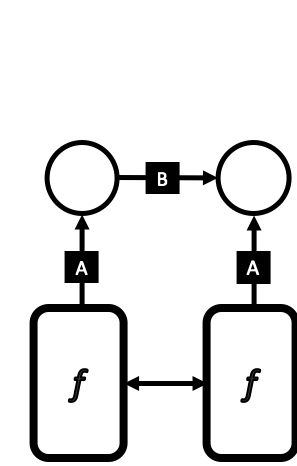} }%
% \hfil%
\subcaptionbox{Multi-Head CRF\label{fig:MHCRF}}[0.22\textwidth]{\includegraphics[width=0.16\textwidth]{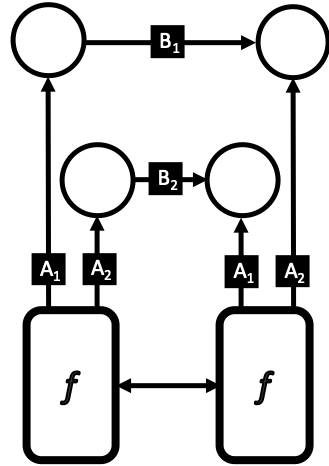}}
% \hfil%
\subcaptionbox{Factorial CRF\label{fig:FCRF}}[0.22\textwidth]{\includegraphics[width=0.16\textwidth]{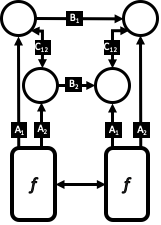}}
% \hfil%
\subcaptionbox{Weighted Factorial CRF\label{fig:WFCRF}}[0.25\textwidth]{\includegraphics[width=0.16\textwidth]{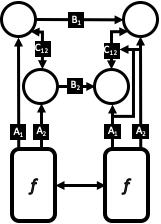}}%
\caption{Different CRF architectures that take as input the output of an LSTM at each time-step in a sequence of two tokens. The circles represent predictions of the labels
at each time-step, and the filled squares represent transformations.}
\end{figure*}    
% \subsection{CRF Architectures}
%ES: trying to avoid repeating "recent work..." sentences.
%Recent work has primarily focused on
%producing an intermediate representation, $\boldsymbol{z}_t$,
%for each token in an input sequence
%for each time-step of a linguistic sequence.
We first define a feature function $\textbf{z}_t = f(\textbf{x}_t,\boldsymbol{\theta})$
as any arbitrary function with parameters $\boldsymbol{\theta}$ that
maps the $t$-th token, $\textbf{x}_t$, in an input text sequence
to a vector representation or \emph{embedding},
$\textbf{z}_t$, 
to facilitate tasks such as sequence labeling.
While feature functions have traditionally been defined by feature engineering, recent state-of-the-art models employ DNNs in the form of CNNs, LSTMs and more recently Transformers \cite{vaswani2017attention} to model sequence based features. As mentioned in Section \ref{sec:related_work}, most sequence labeling approaches feed the output of a feature function $f$ into a basic linear-chain CRF to improve performance. 
In this work, we do not investigate new neural feature functions, 
which may have a task-specific nature.
Instead, we evaluate and extend dynamic CRF models
initially introduced by \citet{sutton2007},
using them to construct new neural architectures for sequence labeling that
 can be applied to arbitrary feature functions.
%evaluate them in a neural setting. 
% For this we are specifically interested in tasks where multiple labels are available for each token. 
We thus denote a feature function, $f$, as an arbitrary neural model 
for sequence tagging and refer to any combination of 
a feature function and a CRF as $f$-CRF.
In the following sections we introduce different
CRF models for multi label sequence tagging,
then in our experiments, we combine
the dynamic CRF models with DNNs for the first time.
    
\subsection{$f$-CRF}

% Many current architectures for sequence labeling leverage a combination of Bidirectional LSTMs \cite{hochreiter1997long} and linear-chain Conditional Random Fields (CRFs) \cite{lafferty2001conditional, sutton2012introduction} first introduced by \cite{huang2015bidirectional}. 
% {\color{red}ES: We refer to any combination of 
% a feature function and a CRF as $f$-CRF.}

Linear-chain CRFs, illustrated in Figure \ref{fig:LCCRF}, model a sequence under the first-order Markov assumption
that the labels are only conditionally dependent on the label of the previous time-step and the features of the current time-step. 
A linear-chain CRF thus factorizes the conditional distribution 
of a sequence of labels given the sequence of tokens
into two main terms:
\begin{equation}
    p(\textbf{y}|\textbf{x}) = \frac{1}{Z(\textbf{x)}} \prod_{t=1}^{T}\exp([\textbf{A}f(\textbf{x}_t;\boldsymbol{\theta})]_{y_t} + \textbf{B}_{y_{t-1},y_t}),
\end{equation}
%TODO: ES: shouldn't it be \boldsymbol{\theta} as it's a vector not a scalar?
where 
%$f()$ denotes the feature function output for time-step $t$ with input token $\textbf{x}_t$.
$\textbf{A}$ is an affine transformation from the output of $f$ to the prediction space,
$y_t$ denotes the index of the label at time-step $t$,
and $\textbf{B}$ is a transition matrix
with entries $\textbf{B}_{a,b}=\ln p(y_t=b | y_{t-1}=a)$ that
define the log probability of the label 
at the current time-step given the label at the previous time-step.
We reduce the notations from $[\textbf{A}f(\textbf{x}_t;\boldsymbol{\theta})]_{y_t}$ to $\textbf{A}\textbf{f}_{t,y}$ for simplicity. %bold f because it would be a vector
$Z(\textbf{x})$ denotes the normalization  factor over all possible sequence states of $\textbf{y}$:
\begin{equation}
    Z(\textbf{x}) = \sum_{\textbf{y} \in \textbf{Y}} \prod_{t=1}^{T}\text{exp}(\textbf{A}\textbf{f}_{t,y} + \textbf{B}_{y_{t-1},y_t}).
\end{equation}

% To simplify the terminology we will refer to linear-chain CRFs as simply CRFs from now on in this paper. 
% TODO: ES: I think you can cut this last sentence as putting the 
% abbreviation in brackets is sufficient.

\subsection{Multi-Head $f$-CRF}
%ES: I have edited the text a bit here
To predict sequence labels for multiple tasks,
we need to adapt the architecture of the $f$-CRF. 
This can be done with a \textit{standard multi-task setup} where the different tasks share the same $f$ function 
with 
separate CRFs as prediction layers for each task.
%ES: avoiding saying "independent" as they are only independent given f; so when you train the complete model, they are not really independent.
%multiple independent CRFs as prediction layers. 
% Different architectures are also possible which have been explored by \citet{ChangpinyoHS18} however these are primarily focused on combining different datasets. The reasoning for the different architectures is primarily based on the assumption that the idosyncracies of the respective datasets require different encodings. In our work we focus only on datasets where multiple labels are provided for sentences and we thus only explore architectures with shared $f$ encodings for each task.   
Our multi-head $f$-CRF architecture, illustrated in Figure \ref{fig:MHCRF}, thus jointly learns 
the weights of a shared $f$ 
(the output of $f$ is the same for each task), 
but learns distinct transition matrices for each task 
$j \in J$:
\begin{equation}
    p(\textbf{y}_j|\textbf{x}) = \frac{1}{Z(\textbf{x)}} \prod_{t=1}^{T}\text{exp}(\textbf{A}_j \textbf{f}_{t,y} + \textbf{B}_{j,y_{t-1},y_t}).  %\forall j \in J
\end{equation}

\subsection{Factorial $f$-CRF}

While multi-head $f$-CRFs jointly learn
the $f$ weights for each of the tasks, they introduce a
conditional
independence assumption between the predicted labels
given $f$. %there is still a dependency through f...
To mitigate this, we revisit factorial CRFs, illustrated in Figure \ref{fig:FCRF},  which are a special case of dynamic CRFs introduced by \citet{sutton2007}. Factorial CRFs model the conditional 
dependency between multiple tasks by introducing the 
% {\color{red}ES: 
log joint probability
%transition 
matrix $\textbf{C}$:
%TODO: ES: can we call this a joint probability matrix or something? It's not really a transition as these are separate tasks and I don't think it is a conditional probability.
\begin{multline}
    p(\textbf{y}_j|\textbf{x}) = \frac{1}{Z(\textbf{x)}} \prod_{t=1}^{T}\text{exp}\Big(\textbf{A}_j \textbf{f}_{t,y} + \textbf{B}_{j,y_{j,t-1},y_{j,t}} + \\ 
    \sum_{\hat{j} \in J \backslash j} \textbf{C}_{j,y_{\hat{j},t},y_{j,t}} \Big), \forall j \in J.
\end{multline} 
This encodes the dependency between tasks at each time-step. Since $\textbf{C}_{j}^T = \textbf{C}_{\hat{j}}$, 
the log joint probability matrix between task $j$ and $\hat{j}$ is shared between the two tasks.
% }
% \begin{equation}
%     \textbf{C}_{\hat{j}}^T = \textbf{C}_{j}.
% \end{equation}

\subsection{Weighted Factorial $f$-CRF}

In practice, 
the labels for the other task, $\hat{j}$, are also uncertain.
Therefore, we enhance factorial CRFs by introducing a 
new variant that weights the matrix $\textbf{C}_{j}$ according to this uncertainty.
%  {\color{red}ES:
 For this, we
scale the log joint probability matrix
by the likelihood of the label for
%$f$ factor of 
the respective other task, $\hat{j}$:
% }
\begin{multline}
    p(\textbf{y}_j|\textbf{x}) = \frac{1}{Z(\textbf{x)}} \prod_{t=1}^{T}\text{exp}(\textbf{A}_j \textbf{f}_{t,y} + \textbf{B}_{j,y_{j,t-1},y_{j,t}} \\ 
    + \sum_{\hat{j} \in J \backslash j} \textbf{A}_{\hat{j}} \textbf{f}_{t,y} \textbf{C}_{j,y_{\hat{j},t},y_{j,t}}). %, \forall j \in J
\end{multline}
This is illustrated in Figure \ref{fig:WFCRF}.

\subsection{Cascaded Weighted Factorial $f$-CRF}

To avoid modeling %interdependent 
dependencies between the labels for all pairs of tasks, 
we specify a cascaded factorial CRF~\cite{sutton2007},
which defines a hierarchy of dependencies.
This decreases the complexity of inference 
as there are no longer circular dependencies between 
the labels for different tasks, meaning we can avoid expensive
loopy dynamic programming \cite{murphy1999loopy}.
% TODO: ES: citation?
% approaches like loopy belief propagation. 
We can define a hierarchical setup by
specifying an ordered list of tasks $J = [1,2,3,...]$,
where each task, $j$, is dependent on $\hat{j}$ iff $j > \hat{j}$. For this type of hierarchical setup, 
cascaded factorial CRFs can be defined as:
\begin{multline}
    p(\textbf{y}_j|\textbf{x}) = \frac{1}{Z(\textbf{x)}} \prod_{t=1}^{T}\text{exp}(\textbf{A}_j \textbf{f}_{t,y} + \textbf{B}_{j,y_{j,t-1},y_{j,t}} + \\ 
    \sum_{\hat{j}=1}^{j-1}%{\hat{j} \in J, j > \hat{j} }
    \textbf{A}_{\hat{j}} \textbf{f}_{t,y} \textbf{C}_{j,y_{\hat{j},t},y_{j,t}}), \forall j \in J.
\end{multline} 
This structure is similar to the weighted factorial depicted in Figure \ref{fig:WFCRF},
except that the connections between tasks are only present for tasks with indices $\hat{j}<j$.

\begin{figure*}
  \centering
  \includegraphics[width= \linewidth]{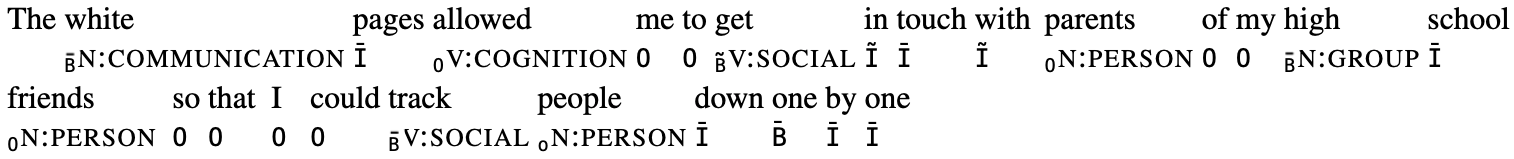}
  \caption{Example Streusle dataset \cite{schneider2015corpus} including the supersenses indicated by $_\text{B}\text{N}$ and $_\text{B}\text{V}$ classes and MWE indicated by the BIO tags. }
  \label{fig:example_streusle}
%   \vspace{}
\end{figure*}

\begin{figure*}[htp]
  \centering
    \small
    \fbox{
    \parbox{0.95\linewidth}{
     \raggedright
    \sethlcolor{green}\hl{
    First I wanted to see if the problem was new, so I checked the teacher's observations
    }. 
    \sethlcolor{yellow}\hl{\mbox{\underline{As it was the same back then}}},
    \hl{I ruled out a }
    \hl{trauma or another dramatic event}. 
    \sethlcolor{cyan}\hl{I was then undecided between autism and ADHD}, \underline{since his social}
    \underline{behaviour seems to be problematic and that's a sign for both diagnoses}. 
    \sethlcolor{yellow}\hl{In the end, I settled on ADHD 
    \mbox{\underline{since his script}} 
    \mbox{\underline{seems chaotic and unorganised}} and \mbox{\underline{because he seems to have some friends despite his difficult behaviour}}}.
    }}
    \caption{Example text from the TEd dataset, with 
    highlighted spans for EG (green), EE (underlined), DC (yellow), HG (blue).}
    \label{fig:famulus_examples}
\end{figure*}

\section{Datasets}

We evaluate the different CRF architectures
%conduct experiments 
on three very diverse datasets in the languages English and German. 
The Streusle dataset \cite{schneider2015corpus} focuses on extracting the semantics of the text, introducing many different labels of supersense categories and identifying multi-word expressions. The MalwareTextDB \cite{lim2017malwaretextdb} on the other hand has the task of extracting malicious entities, providing a very difficult NER task. While the first two datasets are on sentence level and in English, FAMULUS \cite{schulz2019challenges} is a document-level sequence labeling task in German. It focuses on diagnostic reasoning for the medical and teacher education domains and consists of 4 interdependent tasks.

\paragraph{Streusle}
The Streusle dataset \cite{schneider2015corpus} consists of three tasks. POS tagging, supersense categories (SSC) and multi-word expressions (MWE). SSC refers to top-level hypernyms from WordNet \cite{miller1998wordnet}, which are designed to be broad enough to encompass all nouns and verbs \cite{miller1990nouns, fellbaum1990english}. In total the SSC task consists of 26 noun and 15 verb categories. MWEs consist of  single- and multi-word noun and verb expressions with supersenses that encompass idioms, light verb constructions, verb-particle constructions, and compounds \cite{sag2002multiword}. We provide an example annotation of the Streusle dataset in Figure \ref{fig:example_streusle}. The dataset consists of $2,723$ train,  $554$ dev, and $535$ test data points.

\paragraph{Malware}
The MalwareTextDB \cite{lim2017malwaretextdb} consists of 39 annotated Advanced Persistent Threat (APT) reports released by APTnotes\footnote{\href{https://github.com/
aptnotes/}{https://github.com/aptnotes/}}. The dataset is targeted for the cybersecurity domain for automatically detecting malicious entities. In total, the dataset consists of $6,952$ sentences\footnote{\href{https://github.com/juand-r/entity-recognition-datasets}{https://github.com/juand-r/entity-recognition-datasets}} over all 39 domains, which  we split into $4,952$ train, and $1,000$ development and test sets.  
We extend this dataset by an additional task (described below) by using the spacy.io framework\footnote{\href{https://spacy.io/}{https://spacy.io/}}
 to obtain silver part-of-speech tags.

\paragraph{FAMULUS}

\begin{table}[ht]
\small 
\centering
\begin{tabular}{lrlrl}
\toprule
      & \multicolumn{2}{c}{\textbf{Med}} & \multicolumn{2}{c}{\textbf{TEd}} \\
      & \#       & av. len      & \#        & av.len      \\
         \midrule
\textbf{EG/EE} & 5        & 3.8          & 8         & 7.9         \\
\textbf{HG/DC} & 4        & 8.5          & 2         & 22.0        \\
\textbf{DC/EE} & 342      & 9.8          & 143       & 10.9        \\
\textbf{EG/HG} & 0        & -            & 3         & 6.0         \\
\textbf{HG/EE} & 12       & 5.7          & 8         & 11.1        \\
\textbf{EG/DC} & 4        & 6.8          & 3         & 11.7  \\  
\bottomrule
\end{tabular}
\caption{Corpus statistics in terms of absolute number~(\#) and
average number of tokens (av. len), where EE/EG (and similar) denotes an overlap of an EG and EE segment.}
\label{tab:Famulus_statistics}
\end{table}
The FAMULUS datasets \cite{schulz2019challenges, schulz2019analysis} comprise diagnostic reasoning annotations 
in the Medical (Med) and Teacher Education (TEd) domains. 
Each dataset contains summaries written by students of virtual patients (\emph{cases}),
in which the students reason over possible symptomatic diagnoses. % for each domain. 
The argumentative structure of the diagnoses is categorized 
into 
%different classes, titled 
\textit{diagnostic activities} \citep{Fischer2014}, 
covered by sub-spans of the text.  

The dataset consists of 4 diagnostic activity classes: \emph{hypothesis generation} (HG; the derivation of possible answers to the problem), \emph{evidence generation} (EG; the derivation of evidence, e.g., through deductive reasoning or observing phenomena), \emph{evidence evaluation} (EE; the assessment of whether and to which degree evidence supports an answer to the problem), and \emph{drawing conclusions} (DC; the aggregation and weighing up of evidence and knowledge to derive a final answer to the problem),
discussed in detail by \citet{schulz2019challenges}. 
A translated labeled example %from the TEd dataset 
is shown in Figure \ref{fig:famulus_examples}.

While the datasets consist of 4 tasks, only two of them (DC and EE) have many examples of overlapping labels, as can be seen in Table \ref{tab:Famulus_statistics}. This means that while DC and EE are highly dependent on each other, EG and EE are mostly disjoint from the other tasks. The TEd dataset consist of $724$ train, $110$ development and  $110$ test data points. The Med dataset consists of $847$ train, $130$ development, and  $130$ test data points.

% While the diagnostic texts of the two domains are inherently different, 
% the cases within each domain are also disparate, covering different symptomatic nuances. 
% This fundamentally increases the complexity of the task while radically reducing the data size for each case. Table \ref{tab:task1_class_count} exhibits the average number of class labels across the two domains. This emphasises, that as well as the limited dataset size,
% the label distribution is highly skewed (e.g. EE vs. EG). 

\section{Experiments}
\label{sec:Experiments}
In this section we describe our experimental procedures including how we simulate low resource scenarios, our hyper-parameter search as well as our inference strategy. 

\subsection{Low Resource Training Splits} 
We simulate low resource settings for the Streusle and Malware datasets by randomly splitting the training data into smaller sets. We create 4 
training sets consisting of 100, 500, and 1000 random samples and the full dataset respectively. In order to keep the performance comparable, we keep the full development and test set for all scenarios.

\subsection{Feature Function and Hyper-parameters}
As a feature function $f$ for the representation of the time-steps we take the 
commonly-used BiLSTM architecture throughout all our experiments. This architecture has been shown to perform well on various sequence labeling tasks in combination with CRFs \cite{huang2015bidirectional, lample2016neural, sgaard2016, ma2016end, reimers2017, bingel2017identifying, akbik2018,  Schulz2019aaai}. 
As input to the BiLSTM we combine character and pretrained word embedding representations. For the Streusle and Malware datasets we take pretrained Glove embeddings \cite{pennington2014glove}  in English and for Famulus we take pretrained FastText embeddings \cite{bojanowski2017enriching} in German. 

\begin{table}[htp]
\small
\centering
\begin{tabular}{ll}
\toprule
\textbf{Hyperparameter} & \textbf{Values}    \\
\midrule
\# Layers      & $1, 2, 3$   \\
Hidden Size    & $256, 512$  \\
Batch Size     & $8, 16, 32$ \\
               \bottomrule        
\end{tabular}
\caption{Hyper-parameter settings for the BiLSTM feature function $f$ which we randomly sample over for all experiments }
\label{tab:Hyperparameters}
\end{table}
We follow the widely-used training procedure for BiLSTM-CRFs,
where we first compute $f(\textbf{x}_t,\boldsymbol{\theta}), \forall t \in T$ 
using bidirectional LSTMs, 
where the output of the forward and backward LSTM are concatenated for the respective time-steps. 
The inputs to the LSTM at each time-step are embedding representations of the respective 
words and the output of a learned character language model following \citet{akbik2018}. 
Following this, we compute the forward and backward pass to compute the gradients for the 
LSTM parameters, $\boldsymbol{\theta}$, and the matrices $\textbf{A}$, $\textbf{B}$ and 
$\textbf{C}$. 
For more detail we refer to \citet{lafferty2001conditional, sutton2012introduction}. 

For all experimental setups we randomly sample hyper-parameter settings listed in Table \ref{tab:Hyperparameters}. We use Adam \cite{kingma2014adam} for optimization with default settings, however, we perform linear learning rate warm-up over the first epoch. We perform gradient clipping set to $5.0$.

 We follow \citet{reimers2017} by conducting 5 random seed runs for each hyper-parameter setting. We average the results of each run on the development set.  We train all models until convergence on the loss of the development set and perform inference on the development set subsequently. 
 In our results, we report the average test set scores
 for the best average development setting.
 
%  We train all architectures using the default settings of Adam, however have a linear warm-up phase over the first epoch. For inference we use the Viterbi algorithm, which we combine with Loopy-Belief propagation for the factorial CRF experiments. We fix Loopy-Belief to a maximum of 10 loops as we find that it performs on par with higher numbers. 

\subsection{Inference}

For decoding the optimal tags during inference,
we use the dynamic programming algorithm, Viterbi \cite{forney1973viterbi}. Due to the inter-dependency nature of the factorial CRF structure, we require a looping dynamic programming algorithm to infer the most likely label of the dependent tasks. 
For this we run loopy belief propagation \cite{murphy1999loopy} at each time-step between all the tasks for the factorial settings. This is unfortunately greedy in nature with respect to the dependent tasks, meaning
that the algorithm is not guaranteed to find the optimal solution,
but this approach resolves otherwise intractable computation and has been shown to work well in practice~\cite{murphy1999loopy}.

\section{Results}
In this section we report and discuss the results from our three datasets, Streusle, Malware and FAMULUS. 
% The Streusle and Malware datasets contain a POS tagging task, which we consider here as
% an auxiliary task and thus do not report the results. 
We compare the different models introduced in Section \ref{sec:MTST} trained using the setup described in Section \ref{sec:Experiments}. Here we would like to point out that the multi-head (MH) model is the traditional multi-task setup that includes a shared feature function $f$, such as a BiLSTM, but has separate CRFs for each task, such that the inter-dependency is not explicitly modeled.

\subsection{Streusle}

% \begin{table*}[htp]
% \centering
% \begin{tabular}{lllllll}
% \toprule
%  \textit{SSC}     & \textbf{ST}      & \textbf{MH}         & \textbf{Fac} & \textbf{WFac} & \textbf{CFac} & \textbf{WFac (Cl)} \\
%      \midrule
% \textbf{100}  & \textbf{0.3557} & 0.3376     & 0.3265    & 0.3253          & 0.3165             & 0.3232                        \\
% \textbf{500}  & \textbf{0.5073} & 0.4930 & 0.4658   & 0.4762          & 0.4796         & 0.4812                       \\
% \textbf{1000} & \textbf{0.5770} & 0.5666    &  0.5292   & 0.5368         & 0.5523            & 0.5548                       \\
% \textbf{2723} & \textbf{0.6306} & 0.6232    &   0.5821        & 0.6037            & 0.6083            & 0.6214    \\
% \bottomrule
% \end{tabular}
% \caption{F1-Results of the different CRF architectures on the Streusle \cite{schneider2015corpus} SSC task. Single task (ST), multi-head (MH), factorial (Fac), weighted factorial (WFac), cascaded factorial (CFac), weighted factorial with loss clamping (WFac (Cl)) }
% \label{table:results_SSC}
% \end{table*}

\begin{table*}[htp]
\small
\centering
\begin{tabular}{llrrrrr}
\toprule
 \textbf{Task} & \textbf{\# Train}   & \textbf{ST}      & \textbf{MH}      & \textbf{Fac} & \textbf{WFac} & \textbf{CFac} \\ % & \textbf{WFac (Cl)} \\

     \midrule
\multirow{4}{*}{\textit{POS}} &\textbf{100}  & \textbf{ 79.91}  $\pm 0.3 $ & 78.93   $\pm 0.4 $    &  75.72 $\pm 0.2 $   &   78.30     $\pm 0.5 $       &   78.77   $\pm 0.4 $     \\ %       & 77.84  $\pm 0.6 $                      \\
&\textbf{500}  & \textbf{ 88.53} $\pm 0.2 $ &  87.76 $\pm 0.6 $  &  86.43  $\pm 0.5 $  &    87.42  $\pm 0.5 $       &  88.28   $\pm 0.2 $     \\ %   &   87.70   $\pm 0.3$                   \\
&\textbf{1000} & \textbf{91.00 } $\pm 0.2 $ &  90.94 $\pm 0.32 $   &   89.78 $\pm 0.3 $  &   90.21   $\pm 0.4 $      &   \textbf{91.00} $\pm 0.5 $       \\ %     &  90.30  $\pm 0.4 $                     \\
&\textbf{2723} & \textbf{93.25 } $\pm 0.4 $  &  92.91  $\pm 0.3 $      &   91.13   $\pm 0.4 $      &   92.40   $\pm 0.3 $          &  92.87  $\pm 0.4 $      \\ %       &   92.54  $\pm 0.4 $ \\  

     \midrule
\multirow{4}{*}{\textit{SSC}} &\textbf{100}  & \textbf{35.57}  $\pm 0.8$ & 33.76  $\pm 0.7$    & 32.65 $\pm 2.9$   & 32.53     $\pm 1.3$       & 31.65    $\pm 1.8$    \\ %        & 32.32  $\pm 1.4$                      \\
&\textbf{500}  & \textbf{50.73} $\pm 0.7$ & 49.30 $\pm 0.4$  & 46.58  $\pm 0.7$  & 47.62    $\pm 0.6$       & 47.96   $\pm 1.0$  \\ %       & 48.12    $\pm 1.4$                   \\
&\textbf{1000} & \textbf{57.70} $\pm 0.5$ & 56.66 $\pm 0.6$   &  52.92 $\pm 1.1$  & 53.68    $\pm 0.3$      & 55.23  $\pm 0.3$      \\ %      & 55.48  $\pm 0.8$                     \\
&\textbf{2723} & \textbf{63.06} $\pm 0.1$  & 62.32   $\pm 0.7$      &   58.21  $\pm 0.8$      & 60.37    $\pm 0.7$          & 60.83  $\pm 0.4$       \\ %      & 62.14   $\pm 0.7$ \\  
  
     \midrule
\multirow{4}{*}{\textit{MWE}} & \textbf{100}  & 3.23  $\pm 1.5$  & 7.66 $\pm 3.5$ & 15.27 $\pm 2.5$    & \textbf{17.32}  $\pm 3.0$          & 12.03         $\pm 2.2$   \\ %   & 15.76 $\pm 4.0$                      \\
                    & \textbf{500}  & 25.07  $\pm 2.6$  & 32.34 $\pm 1.1$ & 32.71  $\pm 1.4$    & \textbf{39.10} $\pm 3.0$        & 23.91  $\pm 2.7$     \\ %    & 38.58    $\pm 3.0$              \\
                    & \textbf{1000} & 40.00  $\pm 1.2$  & 38.67  $\pm 2.0$&  38.53 $\pm 0.7$   & \textbf{45.05} $\pm 0.5$           & 34.04 $\pm 2.2$          \\ %   & \textbf{45.08} $\pm 0.8$                     \\
                   &   \textbf{2723} & 50.51  $\pm 1.1$  & 50.17 $\pm 0.6$ &  50.51  $\pm 1.3$  & \textbf{51.46}  $\pm 1.2$          & 43.22     $\pm 1.0$        \\ %  & \textbf{52.92} $\pm 0.3$ \\
\bottomrule
\end{tabular}
\caption{F1-Results of the different CRF architectures on the Streusle tasks. Single task (ST), multi-head (MH), factorial (Fac), weighted factorial (WFac), cascaded factorial (CFac).
%, WFac with loss clamping (WFac (Cl)) 
}
\label{table:results_MWE}
\end{table*}

\begin{figure}
  \centering
  \includegraphics[width= \linewidth]{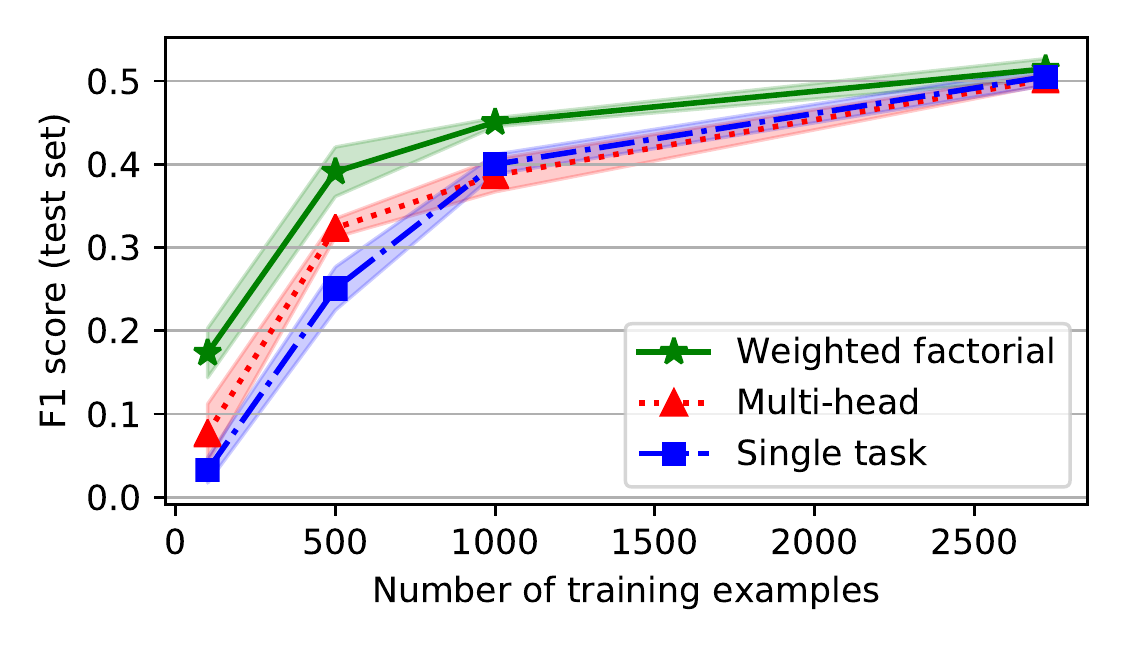}
  \caption{Results %of the different CRF architectures 
  on the MWE task of the Streusle dataset. 
  The lines represent the mean results with bands indicating one standard deviation.
  %The x-axis denotes the number of training examples used whereas on the Y-axis reports the F1-score on the test set.
  }
  \label{fig:Streusel_MWE_results}
%   \vspace{}
\end{figure}

% For the Streusle dataset, we initially train models for 
% the supersense categories (SSC) and multi-word expressions (MWE) tasks 
% separately in the $f$-CRF scenario. 
% In this case the representation of $f$ is not shared between the tasks. For joint representations, we compare multi-head $f$-CRFs, where only the representation $f$ is shared, with different factorial $f$-CRF architectures that model the inter-dependency between tasks. 
The results of the different architectures for \textit{SSC} and \textit{MWE} are presented in Table \ref{table:results_MWE}. 
We can see that sharing representations between the tasks improves performance on the \textit{MWE} task. Especially in the sparse scenario where we only train on 100 instances, the single task setup is outperformed by 14 points. We also find that factorial CRFs outperform the multi-head CRFs over all training data sizes, which is illustrated  in Figure \ref{fig:Streusel_MWE_results}. 
% This shows that gains in performance can be achieved by modeling the inter-dependency between tasks. 

% We further find that our weighted factorial $f$-CRF  adaptation outperforms the traditional factorial $f$-CRF. We argue that this is because this setup takes the label likelihood of the current time-step into account when determining the strength of dependency between tasks. We illustrate the learnt dependency in Figure \ref{fig:Streusle_Heatmap} in which we plot two heatmaps of the joint probability between POS and MWE. The top heatmap shows positive dependencies, for label combinations which are likely to occur. The bottom heatmap shows negative dependencies, thus labels which are very unlikely to occur and thus have learnt to downscale the probability of the two labels appearing for the same token. We also present two anecdotal results in Figure \ref{fig:Pred_Streusle_example}. We find that in many cases the factorial setting is able to predict more correct spans than the Multi-Head and Single-Task setups. In the two anecdotal examples, the factorial model might be able to leverage the fact that \textit{Farell Electric} and \textit{massage therapist} are Nouns which have high probability of being MWEs, and thus leverage the prediction of the POS task to correctly predict the MWE. 

On the other hand, the results of the multi-head as well as factorial settings for the \textit{POS} and \textit{SSC} task are continuously a few points below the single task setting, indicating that there exists interference between tasks as has been reported frequently for multi-task learning \cite{mccloskey1989catastrophic, french1999catastrophic, lee2017fully} . 
However for the harder task \textit{MWE} we see consitent performance gains, indicating that there is a  strong inter-dependency between the  tasks of the Streusle dataset. By explicitly modeling this inter-dependency, large performance gains can be achieved for the \textit{MWE} task. 

% \begin{figure*}
%   \centering
%   \includegraphics[width= 0.6\linewidth]{img/example_streusle.png}
%   \caption{Anecdotal predictions of the different architectures for the MWE task of the Streusle dataset. We can see that the factorial model (WFac) is able to correctly predict the spans where the single task (ST) and the Multi-Head (MH) models fail. }
%   \label{fig:Pred_Streusle_example}
% %   \vspace{}
% \end{figure*}

\subsection{Malware}

\begin{table}[htp]
\small
\centering
\begin{tabular}{lrrr}
\toprule
  \textit{Malw}   & \textbf{ST}      & \textbf{MH}      & \textbf{WFac}    \\
      \midrule
\textbf{100}  & 4.52 $\pm 2.4$    & 15.88 $ \pm 1.0$ & \textbf{16.98} $\pm 1.8$\\
\textbf{500}  & 24.57  $ \pm 1.4$ & \textbf{32.51} $\pm 1.2$ & 29.61 $\pm 0.6$ \\
\textbf{1000} & 36.93  $ \pm 1.9$ & \textbf{38.70} $\pm 1.2$ & 36.70  $\pm 1.3$\\
\textbf{4952} & \textbf{48.25}  $ \pm 1.1$  &  46.94 $\pm 1.4$ & 47.05  $\pm 0.6$\\
\bottomrule
\end{tabular}
\caption{F1-Results of the different CRF architectures on the MalwareTextDB \cite{lim2017malwaretextdb} dataset. Single task (ST), multi-head (MH), weighted factorial (WFac) }
\label{table:results_Malware}
\end{table}

\begin{figure}
  \centering
  \includegraphics[width= \linewidth]{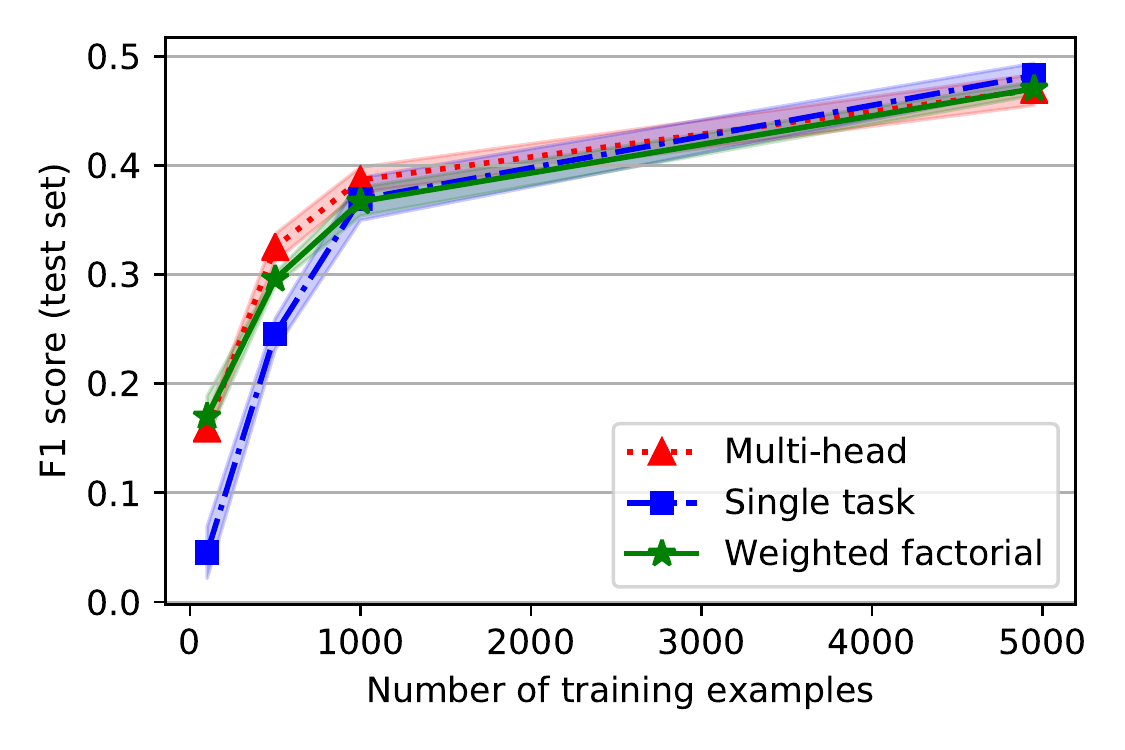}
  \caption{Results %of the different CRF architectures 
  on the Malware dataset. 
    The lines are the mean results, bands indicate one standard deviation.
  %The x-axis denotes the number of training examples used whereas on the Y-axis reports the F1-score on the test set.
  }
  \label{fig:malware_results}
%   \vspace{}
\end{figure}

For the MalwareTextDB dataset we want to probe if it is possible to leverage silver labels of a pretrained part-of-speech tagger\footnote{\href{https://spacy.io/}{https://spacy.io/}} as an auxiliary task to increase performance on the actual task. We find that especially for the  low resource settings where only 100 or 500 training examples exist, both the multi-head $f$-CRF and the factorial $f$-CRF outperform the single task (Table \ref{table:results_Malware}). This indicates that the silver POS labels introduce additional supervision for the task in these settings. However, when the training data increases, the performances of all models are on-par (Figure \ref{fig:malware_results}). This is in line with what can be expected: given sufficient data the sequential representation learning of LSTMs are known to be powerful enough to implicitly learn syntactic features such as POS tags, mitigating the need for explicitly inducing these as labels.  

\subsection{FAMULUS}

\begin{table}[htp]
\small
\centering
\begin{tabular}{lrrr}
\toprule
 \textit{Med}  & \textbf{ST}      & \textbf{MH}      & \textbf{WFac}       \\
 \midrule
\textbf{DC} & 58.63 $ \pm 2.7$  & 59.92 $ \pm 1.4$  & \textbf{62.14}  $ \pm 3.4$   \\
\textbf{EG} & \textbf{71.67} $ \pm 3.5$ & 66.41 $ \pm 3.7$   & 65.25 $ \pm 2.6$ \\
\textbf{EE} & 85.31 $ \pm 0.4$ & 85.80  $ \pm 0.5$ & \textbf{85.89} $ \pm 0.5$   \\
\textbf{HG} & \textbf{59.05} $ \pm 2.1$  & 54.93 $ \pm 3.7$ & 56.56  $ \pm 4.7$  \\
\bottomrule
\end{tabular}
\caption{F1-Results of the different CRF architectures on the Med FAMULUS \cite{Schulz2019aaai} dataset. Single task (ST), multi-head (MH), weighted factorial (WFac) }
\label{table:results_Med}
\end{table}

\begin{table}[htp]
\small
\centering
\begin{tabular}{llll}
\toprule
 \textit{TEd}   & \textbf{ST}     & \textbf{MH}      & \textbf{WFac}    \\
 \midrule
\textbf{DC} & 50.15 $ \pm 7.0$ & 53.57 $ \pm 4.5$  &\textbf{ 54.28}$ \pm 5.0$ \\
\textbf{EG} & \textbf{76.57}  $ \pm 3.3$ & 74.49 $ \pm 1.7$   & 74.44 $ \pm 2.9$\\
\textbf{EE} & 84.09 $ \pm 0.5$ & 85.07  $ \pm 0.7$ & \textbf{85.33} $ \pm 0.9$ \\
\textbf{HG} & \textbf{42.96} $ \pm 2.6$ & 38.89  $ \pm 4.7$   & 36.05 $ \pm 9.5$\\
\bottomrule
\end{tabular}
\caption{F1-Results of the different CRF architectures on the TEd FAMULUS \cite{Schulz2019aaai} dataset. Single task (ST), multi-head (MH), weighted factorial (WFac) }
\label{table:results_TEd}
\end{table}

In Table \ref{table:results_Med} and  \ref{table:results_TEd} we present the results for the different architecture setups for the FAMULUS Med and TEd datasets respectively.  In line with the overlapping labels for each task presented in Table \ref{tab:Famulus_statistics}, we find that the shared representation models (multi-head and factorial) outperform the single task models for \textit{DC} and \textit{EE}. However, for \textit{EG} and \textit{HG}, the single task models are better than the joint representations. This is in line with what can be expected as \textit{EG} and \textit{HG} have almost no overlapping labels with the other tasks. Modeling inter-dependency between the tasks thus hurts performance as the model tries to find a joint representation between the tasks.  For the dependent tasks \textit{DC} and \textit{EE},
we find that the weighted factorial model outperforms the multi-head model for both tasks, 
but with a larger margin for \textit{DC} for both Med and TEd. This indicates that modeling the inter-dependency between the tasks helps the model generalize better by leveraging the prediction of the respective other task. 

\section{Discussion}

%visualisation of matrix C
% rows are BI labels for two classes (see the example)
% the matrix is just split into two for visualisation: at the top, if it's not black it's positive; at the bottom, if it's not white it's negative. 
% the values are values of C, which may be some kind of unnormalised log joint probabilities...

\begin{figure}
  \centering
  \includegraphics[width= \linewidth]{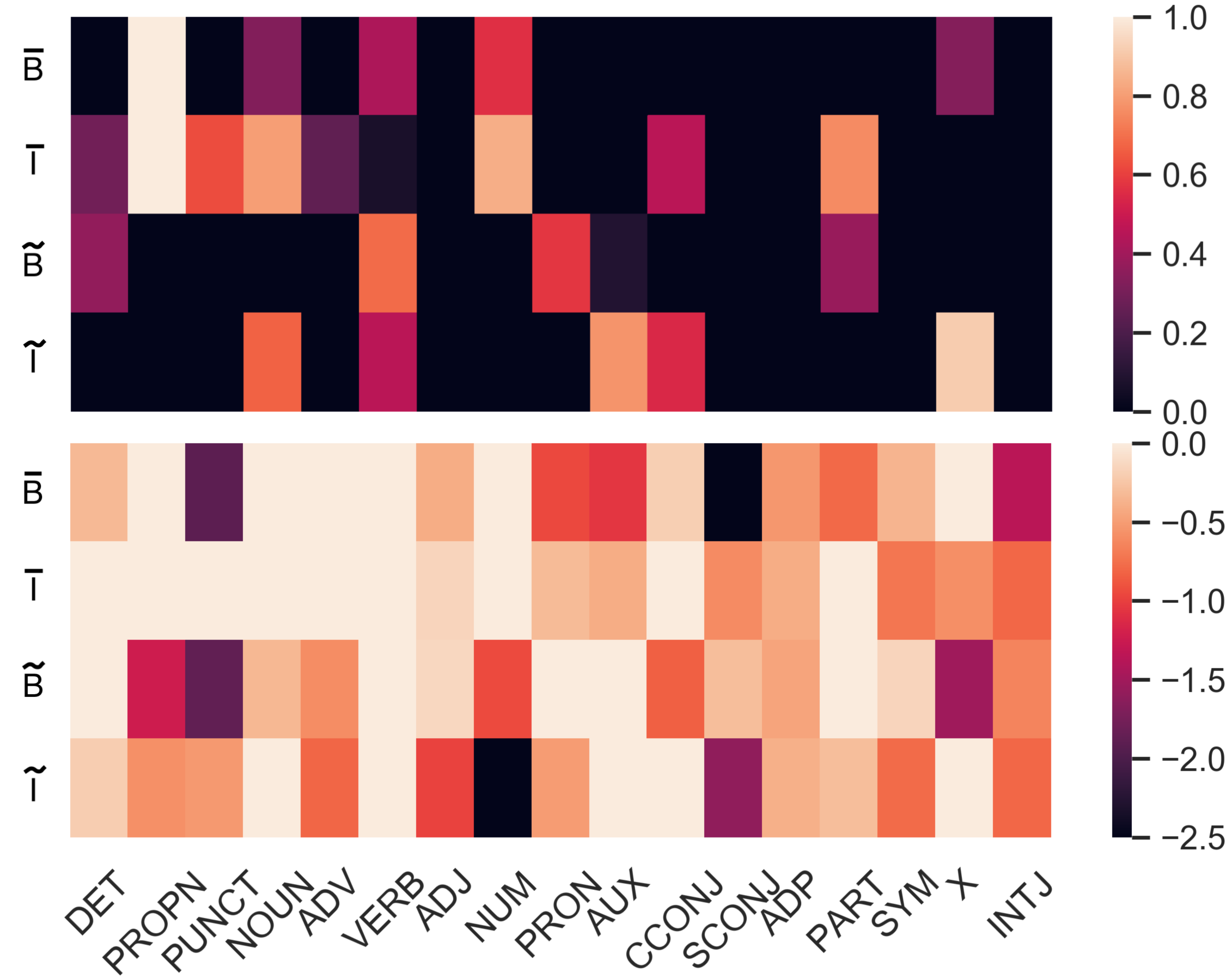}
  \caption{Heatmaps of the joint probability matrix between POS and MWE of the Streusle dataset. The top heatmap shows 
  positive correlations between tasks as non-black entries. 
  The bottom heatmap represents combinations that are unlikely to occur together, thus the values are negative.
  Rows correspond to the beginning and inside token labels for 
  two types of MWE spans, `-' and `$\sim$'. Columns correspond
  to POS tags.}
  \label{fig:Streusle_Heatmap}
%   \vspace{}
\end{figure}

% In our experiments, we have analyzed different settings of 

In our experiments, we %the previous section we presented the results of the different tasks and have 
found that by explicitly modeling the task inter-dependencies, performance gains can be achieved  for many scenarios. This effect can be seen especially in the low-resource settings, in which the weighted factorial model (WFac) outperforms the single task (ST) as well as the traditional multi-task (MH) models. This is also true for cases where we make use of
cheap silver labels from pretrained POS taggers. 

The strongest performance gains can be achieved when combining multiple related tasks with spans that appear infrequently in the dataset, such as \textit{SSC} and \textit{MWE} in the Streusle dataset. There is a strong inter-dependency between the tasks which the model is not able to implicitly learn in the multi-head or single task setting, compared to the explicit dependency representation of WFac.  

An example of the explicit dependencies modeled by WFac is illustrated in Figure \ref{fig:Streusle_Heatmap}.
Here, we plot two heatmaps of the $\textbf{C}$ matrices
that encode the dependencies 
%which contain unnormalized log joint probability functions.
between each of the POS tags and
the MWE labels
for the Streusle dataset
% \footnote{We do not present the heatmap for the \textit{SSC} task due to space limitations, as \textit{SSC} includes 41 classes which would mean that we would plot a 82x8 heatmap.}
. 
The top heatmap shows positive correlations, 
i.e., the label combinations that are likely to occur. 
The bottom heatmap shows negative dependencies,
where labels are unlikely to coincide. 
The heatmaps show that the model has learned  
dependencies between the MWE labels and specific POS tags.
It uses these values to downscale 
the probability of labels which are unlikely to co-occur but upscale those which are likely to appear at the same token.  
%????
By modeling the dependencies explicitly, WFac can
directly leverage the predictions of other tasks.
%, making it easier to learn the dependency. 
In contrast, multi-head models only share the feature function $f$, so require more data to learn to encode the dependency within the deep $f$ model.

While we consistently see performance gains for the multi-task approaches for a subset of the tasks, the performances for other tasks simultaneously deteriorate. For the Streusle dataset we observe gains for \textit{MWE} across all multi-task settings over the single task setting, however for the two other tasks \textit{POS} and \textit{SSC} the single task setting performs the best. Similarly, for the FAMULUS dataset, tasks \textit{EG} and \textit{HG} perform the best in the single-task setting as these do not have many overlapping labels with the respective other tasks. Similar observations of interference between tasks for multi-task learning have been reported frequently in literature \cite{mccloskey1989catastrophic, french1999catastrophic, lee2017fully},  indicating that sharing the entirety of parameters can be harmful for performance for a subset of the tasks. However, when leveraging additional labels for auxiliary tasks, such as the silver POS tags for the Malware dataset, performance drops on the auxiliary tasks can be disregarded as the performance gain on the target task is the objective.

\section{Conclusion}
In this paper, we investigated multi-task sequence tagging, 
introducing neural factorial CRF models that explicitly model the inter-dependencies between different task labels.
We compared different methods for datasets where
multiple labels are available for each example, including single task learning, standard multi-task learning, and factorial CRFs, 
finding strong performance for factorial models in low resource settings where spans of different tasks coincide. 

Similar to what has been reported in literature, we observe interference between tasks in multi-task learning settings, indicating that sharing the entirety of parameters decreases performance on a subset of the tasks. In the future we will investigate recent Adapter approaches \cite{rebuffi2017learning, houlsby2019parameter} which train new parameters within each layer of pre-trained models, to combine them for multi-task learning as proposed by  \citet{pfeiffer20adapterfusion}.

Based on our results, 
we believe that modeling the inter-dependencies between tasks  
could be beneficial during the early stages of dataset creation, 
where only small amounts of data are available. 
Employing such models in a bootstrapping setup to provide annotators with
label suggestions can increase the speed of dataset creation as well as improve the 
inter-annotator agreement~\cite{schulz2019analysis, pfeiffer19}.

\section*{Acknowledgments}
This work has been supported by the German Federal Ministry of Education and Research (BMBF) under the reference 16DHL1041 (FAMULUS). 

\bibliography{acl2019}
\bibliographystyle{acl_natbib}

\end{document}